\documentclass[11pt]{article}

\usepackage[final]{acl}

\usepackage{times}
\usepackage{latexsym}

\usepackage[T1]{fontenc}

\usepackage[utf8]{inputenc}

\usepackage{microtype}

\usepackage{inconsolata}

\usepackage{graphicx}

\usepackage{amsmath}
\usepackage{amssymb}
\usepackage{xspace}
\usepackage[capitalize,noabbrev]{cleveref}

\usepackage{booktabs}
\usepackage{multirow}
\usepackage{makecell}
\usepackage{subfig}
\usepackage[table]{xcolor}
\usepackage{array}
\usepackage{colortbl}
\usepackage{tikz}
\usepackage{pgfplots}

\usepackage{pifont}

\usepackage{algorithm}
\usepackage{algorithmic}

\newcommand{\OURS}{\textsc{DynamixSFT}\xspace}
\newcommand{\papertitle}{\OURS: Dynamic Mixture Optimization\\of Instruction Tuning Collections}

\newcommand{\fullBaseline}{Full Coverage\xspace}
\newcommand{\uniformBaseline}{Uniform Sampling\xspace}

\newcommand{\tulu}{\textsc{Tülu}\xspace}

\newcommand{\mixtureTwo}{\tulu-2-mixture\xspace}
\newcommand{\mixtureThree}{\tulu-3-mixture\xspace}

\newcommand{\priorScaled}{Prior-scaled Boltzmann Exploration\xspace}
\newcommand{\lookaheadReward}{1-Step Look-ahead Reward\xspace}

\newcommand{\cmark}{\ding{51}}
\newcommand{\xmark}{\ding{55}}

\definecolor{First}{HTML}{ea9999}
\definecolor{Second}{HTML}{f4cccc}
\definecolor{Third}{HTML}{F7E6E6}

\title{\papertitle}

\author{
    Haebin Shin$^{1,2,3}${\thanks{\* Work performed at KAIST AI and internship at Microsoft Research.}}
        \quad
    Lei Ji$^2$
        \quad
    Xiao Liu$^2$
        \quad
    Zhiwei Yu
        \\
    \textbf{Hyunwoo Yoo}$^4$
        \quad
    \textbf{Qi Chen}$^2$
        \quad
    \textbf{Yeyun Gong}$^2$  
    \\[1.5ex]
    $^1$University of Michigan
        \quad
    $^2$Microsoft Research
        \quad
    $^3$KAIST AI
        \quad
    $^4$Drexel University
    \\[0.5ex]
    \texttt{haebin@umich.edu}
}

\begin{document}
\maketitle
\begin{abstract}
As numerous instruction-tuning datasets continue to emerge, dynamically balancing and optimizing their mixtures has become a critical challenge.
To address this, we propose \OURS, a dynamic and automated method for instruction-tuning dataset mixture optimization.
We formulate the problem as a multi-armed bandit setup and introduce a \textit{\priorScaled} that softly anchors the updated sampling distribution to the original dataset proportions, thereby preserving the inherent diversity and coverage of the collection.
Sampling probabilities are updated using a lightweight \textit{\lookaheadReward}, reflecting how much the dataset contributes to improving the model’s performance at its current state.
We demonstrate that \OURS effectively optimizes the \mixtureTwo and \mixtureThree collections across 10 benchmarks, while introducing minimal computational overhead over naive sampling.
Furthermore, we provide a comprehensive analysis and visualizations to offer deeper insights into the adaptive dynamics of our method.

\end{abstract}

\section{Introduction}
\label{sec:introduction}




\begin{figure}[t!]
  \centering
  \centerline{\includegraphics[width=1.0\linewidth]{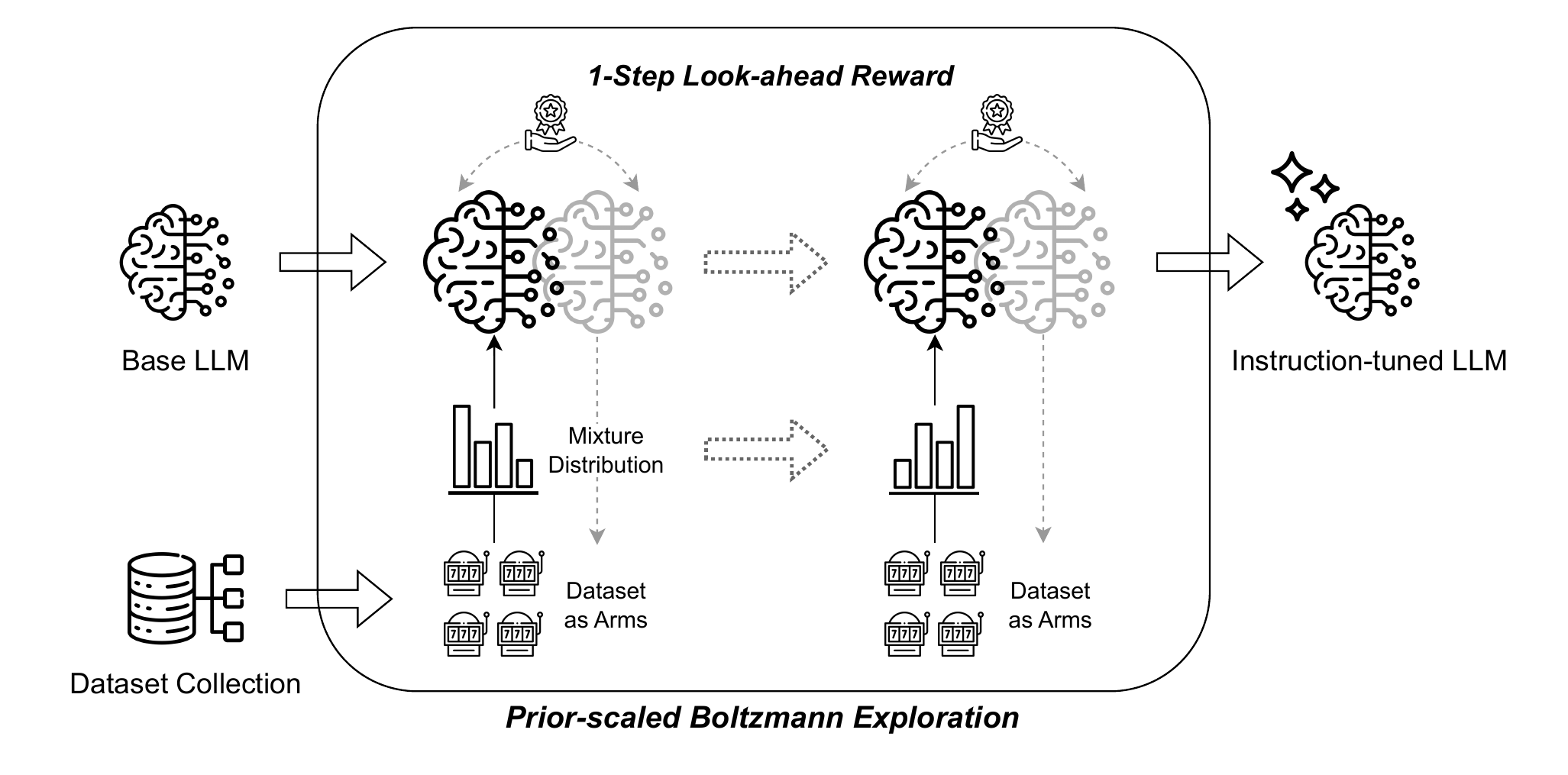}}

  \caption{Overview of \OURS. Given a large collection of instruction-tuning datasets, \OURS treats each dataset as an arm in a \textit{Multi-Armed Bandit} setup. The sampling policy dynamically evolves through \textit{Prior-scaled Boltzmann Exploration}, periodically updated by a lightweight \textit{1-Step Look-ahead Reward} that reflects the model’s current training dynamics, enabling adaptive mixture optimization.}
  \label{fig:main_illustration}
\end{figure}



The mixture of diverse datasets has emerged as a pivotal factor in the post-training stage of large language models (LLMs), significantly enhancing their ability to generalize across a broad spectrum of domains—including instruction following, reasoning, mathematics, coding, and knowledge-intensive tasks~\citep{wei2022finetuned, longpre2023flancollectiondesigningdata}. A well-composed dataset mixture allows LLMs to acquire a broad set of capabilities from heterogeneous sources while preserving robustness and generality. This has led to the development of large-scale mixtures~\citep{NEURIPS2023_ec641387, ivison2023camelschangingclimateenhancing, longpre2023flancollectiondesigningdata,lambert2025tulu3pushingfrontiers}, which offer standardized recipes for instruction tuning with a diverse set of supervision signals. 

However, most curated mixtures are statically defined through manual heuristics or fixed rules~\citep{NEURIPS2023_ec641387, ivison2023camelschangingclimateenhancing, lambert2025tulu3pushingfrontiers}, requiring extensive human effort to design effective mixing strategies.
Several studies rely on additional models~\citep{xie2023doremioptimizingdatamixtures, fan2024dogedomainreweightinggeneralization, liu2025regmixdatamixtureregression, wu-etal-2024-mixture-skills,wang2025hbohierarchicalbalancingoptimization} or resources~\citep{wang-etal-2020-balancing, wu-etal-2021-uncertainty} to guide dynamic dataset mixture optimization.
This highlights the need for a lightweight, self-evolving method that can adapt the dataset mixture during training without additional computation.

To address this challenge, we propose \OURS, a dynamic framework that formulates the allocation of training samples across $k$ heterogeneous data sources as a multi-armed bandit (MAB) problem, where each dataset is treated as an arm. 
To support adaptive yet balanced mixture optimization, we further introduce a \textbf{\priorScaled}, a strategy that softly anchors the updated sampling distribution to the original dataset proportions. This encourages adaptive sampling while preserving the inherent diversity and coverage of the dataset collection.
Additionally, we introduce a lightweight \textbf{\lookaheadReward} to estimate each dataset’s contribution to current training dynamics, enabling the model to prioritize more beneficial data sources based on its own learning progress.

We apply \OURS to the \mixtureTwo~\citep{ivison2023camelschangingclimateenhancing} and \mixtureThree~\citep{lambert2025tulu3pushingfrontiers}, comprising 16 and 19 instruction-tuning datasets, respectively.
We observe consistent gains on LLaMA3.2 1B~\citep{llama3.2} and LLaMA3.1 8B~\citep{grattafiori2024llama3herdmodels} across 10 benchmarks covering knowledge, reasoning, mathematics, coding, and instruction following. 

To better understand the benefits of \OURS, we conduct a comprehensive analysis, including visualizations of the evolving mixture proportions during training. 
This analysis provides practical insights into how the dataset mixture adapts over time in response to the model’s training dynamics, and highlights the role of various factors from the perspective of the bandit algorithm.
Ultimately, our approach enables LLMs to optimize their own dataset mixtures during post-training without the need for hand-crafted tuning, making it feasible to leverage large-scale dataset collections in a more principled and automated manner.

Our contributions are summarized in three folds:
\begin{itemize}
    \item We propose \OURS, a dynamic dataset mixture optimization method for LLM post-training. \OURS formulates the dataset mixture problem as a multi-armed bandit setup and introduces a \textbf{\textit{\priorScaled}} policy that softly anchors the updated sampling distribution to the original dataset proportions. To update sampling probabilities, we further introduce a \textbf{\textit{\lookaheadReward}}, which reflects the model’s current training dynamics and enables adaptive sampling over time.
    \item We apply \OURS to large-scale instruction-tuning collections containing up to \textbf{19 datasets}, evaluating performance across \textbf{10 benchmarks} spanning knowledge, reasoning, mathematics, coding, and general instruction-following. Our method outperforms static and dynamic mixture baselines achieving consistent improvement on both 1B and 8B models with minimal computational overhead.
    \item We present detailed \textbf{analysis and visualizations} of how mixture dynamics change throughout training. This provides insight into the model’s evolving needs and demonstrates how the bandit-based adaptation contributes to effective and interpretable mixture scheduling.
\end{itemize}

\section{Related Works}
\label{sec:related_works}
\subsection{Dataset Mixture for LLMs}
The composition of instruction-tuning data mixtures critically affects model effectiveness and generalization during post-training. The \tulu collection~\citep{NEURIPS2023_ec641387, ivison2023camelschangingclimateenhancing, lambert2025tulu3pushingfrontiers} proposes this mixture by incorporating diverse high-quality datasets with unified formatting and expanded task coverage, but still relies on static or manually curated mixture strategies. We focus on dynamically optimizing the \mixtureTwo~\citep{ivison2023camelschangingclimateenhancing} and \mixtureThree~\citep{lambert2025tulu3pushingfrontiers}.

\subsection{Mixture Optimization}
Despite its importance, dataset mixture optimization remains challenging. Existing approaches are often limited in scope or flexibility. Some are confined to a small number of manually chosen dataset types~\citep{wang-etal-2024-demystifying} or operate at the regularization level rather than directly modifying sampling policies~\citep{xiao2025sftmixelevatinglanguagemodel}. Others require pre-defined supervision from validation sets~\citep{wang-etal-2020-balancing, wu-etal-2021-uncertainty}, introduce additional parameters through proxy models~\citep{xie2023doremioptimizingdatamixtures, pmlr-v235-fan24e, liu2025regmixdatamixtureregression} or actor networks~\citep{wu-etal-2024-mixture-skills, wang2025hbohierarchicalbalancingoptimization}, rely on MoE architectures~\citep{zhu-etal-2025-dynamic}, lack generality across diverse tasks~\citep{chen2025selfevolvingcurriculumllmreasoning, albalak2023efficientonlinedatamixing}. In contrast, \OURS targets large-scale, uncurated instruction-tuning collections covering diverse tasks without external supervision.

\section{Preliminary: Does Data Mixture Matter?}
\label{sec:preliminary}

\newcolumntype{h}{>{\columncolor{black!5}\color{black!50}}c} 

\begin{table*}[t!]
  \centering
  \resizebox{\textwidth}{!}{%
  \begin{tabular}{lcccccccccc|hcc}
    \toprule
    \multicolumn{1}{c}{\multirow{2.5}{*}{\textbf{\makecell{Dataset}}}} &
    \multicolumn{3}{c}{\textbf{General Knowledge}} &
    \multicolumn{2}{c}{\textbf{Reasoning}} &
    \multicolumn{2}{c}{\textbf{Coding}} &
    \multicolumn{2}{c}{\textbf{Mathematical}} &
    \multirow{2.5}{*}{\makecell{IFEval}} &
    \multicolumn{3}{c}{\textbf{AVG}} \\
    
    \cmidrule(lr){2-4}\cmidrule(lr){5-6}\cmidrule(lr){7-8}\cmidrule(lr){9-10}\cmidrule(lr){12-14}
     & MMLU & TQA & PopQA & BBH & DROP & CHE & CHE$+$ & GSM8K & MATH &  & Total & Know+Rea & Code+Math \\ \midrule
    G1                   & \cellcolor{First}29.87 & \cellcolor{First}41.77 & \cellcolor{First}16.35 & \cellcolor{First}32.56 & \cellcolor{Third}27.37 & \cellcolor{Third}36.03 & \cellcolor{Second}31.83 & \cellcolor{Second}9.62 & \cellcolor{Third}2.38 & \cellcolor{Third}22.73 & 25.05 & \cellcolor{First}29.58 & \cellcolor{Second}19.97 \\
    G2                   & \cellcolor{Second}29.58 & \cellcolor{Second}40.61 & \cellcolor{Second}16.07 & \cellcolor{Third}31.85 & \cellcolor{Second}27.57 & \cellcolor{Second}38.68 & \cellcolor{Third}30.77 & \cellcolor{Third}6.82 & \cellcolor{First}3.44 & \cellcolor{First}25.34 & 25.07 & \cellcolor{Second}29.14 & \cellcolor{Third}19.93 \\
    G3                   & \cellcolor{Third}28.68 & \cellcolor{Third}38.31 & \cellcolor{Third}15.84 & \cellcolor{Second}31.88 & \cellcolor{First}27.62 & \cellcolor{First}38.93 & \cellcolor{First}32.26 & \cellcolor{First}14.93 & \cellcolor{Second}2.90 & \cellcolor{Second}24.58 & 25.59 & \cellcolor{Third}28.47 & \cellcolor{First}22.26 \\
    \arrayrulecolor{black!70}\midrule
    G1 + G2                   & \cellcolor{First}31.54 & \cellcolor{First}41.73 & \cellcolor{First}16.70 & \cellcolor{First}32.71 & \cellcolor{Third}28.00 & \cellcolor{Third}36.84 & \cellcolor{Third}31.71 & \cellcolor{Third}8.18 & \cellcolor{Third}2.75 & \cellcolor{Third}25.32 & 25.55 & \cellcolor{First}30.14 & \cellcolor{Third}19.87 \\
    G2 + G3                   & \cellcolor{Third}30.72 & \cellcolor{Third}39.99 & \cellcolor{Third}16.31 & \cellcolor{Third}32.28 & \cellcolor{First}28.20 & \cellcolor{First}41.32 & \cellcolor{First}33.53 & \cellcolor{First}14.10 & \cellcolor{First}2.99 & \cellcolor{First}28.09 & 26.75 & \cellcolor{Third}29.50 & \cellcolor{First}22.99 \\
    \arrayrulecolor{black!70}\midrule
    G3 $\rightarrow$ G2 $\rightarrow$ G1                  & \cellcolor{First}30.03 & \cellcolor{Third}40.40 & \cellcolor{Third}16.40 & \cellcolor{First}33.26 & \cellcolor{Third}28.34 & \cellcolor{Third}38.34 & \cellcolor{Third}31.66 & \cellcolor{Third}12.13 & \cellcolor{Third}2.28 & \cellcolor{Third}25.69 & 25.85 & \cellcolor{First}29.69 & \cellcolor{Third}21.10 \\
    G1 $\rightarrow$ G2 $\rightarrow$ G3                  & \cellcolor{Third}29.35 & \cellcolor{First}40.94 & \cellcolor{First}16.74 & \cellcolor{Third}32.89 & \cellcolor{First}27.57 & \cellcolor{First}38.78 & \cellcolor{First}32.25 & \cellcolor{First}14.25 & \cellcolor{First}3.34 & \cellcolor{First}28.09 & 26.42 & \cellcolor{Third}29.50 & \cellcolor{First}22.16 \\
    \arrayrulecolor{black}\bottomrule
  \end{tabular}
  }
  \caption{Performance trends across different data mixtures and training curricula. Different mixture groups favor different capabilities: G1 improves knowledge and reasoning, while G3 strengthens coding and math.}
  \label{tab:preliminary-results}
\end{table*}




Our hypothesis is that even when training on the same amount of data, different combinations can lead to qualitatively different learning behaviors. To explore the effect of their mixture and ordering, we conduct a simple pilot study using the \mixtureTwo on LLaMA3.2 1B~\citep{llama3.2}. We split the given collection into three equal groups (each sub-dataset is independently divided into three groups and then merged), according to token-averaged negative log-likelihood (NLL) scored by Qwen3-30B-A3B~\citep{yang2025qwen3technicalreport}: high-loss (G1), medium-loss (G2), and low-loss (G3). In~\Cref{tab:preliminary-results}, training on G1 tends to improve general knowledge and reasoning benchmarks, while G3 shows stronger gains on coding and math tasks. Moreover, simply changing the curriculum order across the same partitions leads to different trends in overall performance.

These results suggest that even within a heterogeneous collection, there may exist state-dependent optimal mixtures rather than a single static recipe. This motivates us to dynamically adjust sampling during training to optimize the data mixture based on the model's evolving needs.

\section{Dynamic Mixture Optimization}
\label{sec:method}

\begin{algorithm}[t]
\caption{\OURS}
\label{alg:method}
\begin{algorithmic}[1]
\REQUIRE SFT Dataset Collection $\mathcal{C} = \{\mathcal{D}_1, \dots, \mathcal{D}_K\}$; EMA values $Q \in \mathbb{R}^K$; Prior probabilities $p^{(0)} \in \Delta_K$; Sharpness factor $\beta$; Uniformity factor $\gamma$; EMA smoothing factor $\alpha$; Total steps $T$; Update interval $T_{\text{update}}$; Batch size $B$
\STATE Initialize $Q_k \gets 0$ for all $k \in \{1, \dots, K\}$
\FOR{step $t = 1$ to $T$}
    \STATE $B_t \gets \emptyset$
    \WHILE{$|B_t| < B$}
        \STATE Compute Multi-Armed Bandit probs:
        \[
        p_k \gets (1 - \gamma) \cdot \frac{\exp(\beta \cdot Q_k) \cdot p^{(0)}_k}{\sum_j \exp(\beta \cdot Q_j) \cdot p^{(0)}_j} + \gamma / K
        \] 
        \STATE Sample dataset $\mathcal{D} \sim p$
        \STATE Sample example $x \sim \mathcal{D}$ uniformly
        \STATE Add to batch: $B_t \gets B_t \cup \{x\}$
    \ENDWHILE
    \STATE Train model with batch $B_t$
    \IF{$t \mod T_{\text{update}} = 0$}
        \FOR{each $\mathcal{D}_k \in \mathcal{C}$}
            \STATE Sample next batch $B_{\mathcal{D}_k}$ from $\mathcal{D}_k$
            \STATE Compute reward:
            \[
            r_{k} \gets \frac{1}{|B_{\mathcal{D}_k}|} \sum_{x \in B_{\mathcal{D}_k}} \frac{\mathcal{L}_{\text{pre}}(x) - \mathcal{L}_{\text{post}}(x)}{\mathcal{L}_{\text{pre}}(x) + \epsilon}
            \]
            \STATE Update EMA: 
            \item[] \( 
            \begin{aligned}
            Q_{k} \gets \alpha \cdot Q_{k} + (1-\alpha) \cdot r_{k}
            \end{aligned}
            \)
        \ENDFOR
    \ENDIF
\ENDFOR
\end{algorithmic}
\end{algorithm}

We introduce \OURS, a method that dynamically adjusts dataset mixtures during training~\Cref{alg:method}.

\subsection{Mixture as a Multi-Armed Bandit}
Considering the nature of large-scaled instruction-tuning dataset collections~\citep{NEURIPS2023_ec641387, ivison2023camelschangingclimateenhancing, lambert2025tulu3pushingfrontiers}, we formulate the problem of sampling an optimal batch from $K$ candidate datasets as a multi-armed bandit (MAB) problem.
To quickly adapt to the non-stationary reward distribution that naturally arises during fine-tuning, we adopt a Boltzmann exploration strategy~\citep{sutton1998reinforcement, NIPS2017_b299ad86, gupta2019boltzmannexploration}. A typical Boltzmann exploration~\citep{sutton1998reinforcement} computes sampling probabilities using the following $\operatorname{softmax}(\beta \cdot Q)$:
\begin{equation}
\label{eq:softmax}
    \frac{\exp(\beta \cdot Q_k)}{\sum_j \exp(\beta \cdot Q_j)},
\end{equation}
which treats all reward values $Q$ as equally interpretable across datasets.

However, treating all rewards as equally comparable across datasets can skew the sampling distribution toward specific datasets, failing to account for their inherent characteristics.
In the SFT phase, both \textit{diversity} and \textit{coverage} of the training instances are critical. Notably, we assume that the original dataset proportions—often based on dataset size—implicitly reflect each dataset’s inherent characteristics and its intended contribution to domain coverage. For example, LIMA~\citep{NEURIPS2023_ac662d74} is small but well-curated, whereas FLAN~\citep{longpre2023flancollectiondesigningdata} is relatively large to cover a broader range of domains.

To account for this, we introduce a \textbf{\priorScaled} that explicitly incorporates the underlying dataset distribution $p^{(0)}$ as a prior, extending Equation~\ref{eq:softmax} as follows:
\begin{equation}
\label{eq:prior_scaled}
    \frac{\exp(\beta \cdot Q_k) \cdot p^{(0)}_k}{\sum_j \exp(\beta \cdot Q_j) \cdot p^{(0)}_j}.
\end{equation}
Our prior-scaled policy softly anchors the learned sampling distribution to the original proportions, allowing for adaptive reward-driven updates while preserving the broad coverage and diversity encoded in the original mixture.

Nonetheless, a never-sampling problem may still arise due to the non-stationary nature of reward dynamics. To mitigate this, we apply a \textbf{Minimum Floor Probability} ($\gamma / K$), ensuring that each dataset retains a non-zero probability of being sampled. This guarantees uniform exploration and facilitates adaptation to sudden shifts in reward dynamics.


As a result, we formulate the final bandit sampling probability as shown in~\Cref{eq:bandit_probs}. 
Here, the sharpness parameter $\beta$ controls the exploitation strength of the Prior-Scaled Boltzmann term, while the uniformity factor $\gamma$ provides the exploration guarantee through the $\gamma/K$ term, striking an effective balance between exploitation and exploration.

\begin{equation}
\label{eq:bandit_probs}
    (1 - \gamma) \cdot \frac{\exp(\beta \cdot Q_k) \cdot p^{(0)}_k}{\sum_j \exp(\beta \cdot Q_j) \cdot p^{(0)}_j} + \gamma / K
\end{equation}

%



\subsection{\lookaheadReward}
The goal of updating the bandit probabilities is, intuitively, to assign higher sampling probability to datasets that are expected to contribute more effectively to the model's current training progress.
Concretely, at each update interval $T_\text{update}$, we estimate this contribution by performing a \textbf{1-Step Look-ahead $\Delta$-Loss}: for each dataset, we temporarily update the model by one gradient step on a mini-batch, and then measure how much the model loss decreases as a result. This provides an immediate signal of each dataset’s utility without requiring additional supervision or permanent parameter updates. This 1-step look-ahead mechanism not only captures immediate learning signals but also serves as a proxy for long-term convergence, enabling state-dependent sampling that aligns with the model's evolving needs.
We further provide theoretical support for the long-term effects of immediate reward signals in~\Cref{appendix:theoretical_support}.

The reward for dataset $D_i$ is computed as follows:
\begin{equation}
    r_{\mathcal{D}_i} \gets \frac{1}{|B_{D_i}|} \sum_{x \in B_{D_i}} \frac{\mathcal{L}_{\text{pre}}(x) - \mathcal{L}_{\text{post}}(x)}{\mathcal{L}_{\text{pre}}(x) + \epsilon}
\end{equation}
Here, $\mathcal{L}_{\text{pre}}(x)$ and $\mathcal{L}_{\text{post}}(x)$ denote the loss before and after the virtual one-step update, respectively, and 
$\epsilon$ is a small constant for numerical stability.
We denote by $B_{D_i}$ a mini-batch sampled from dataset $D_i$ for reward estimation.
To mitigate scale differences and occasional negative rewards, we normalize reward across datasets using min–max normalization.
This lightweight look-ahead reward encourages the bandit scheduler to dynamically favor data sources that yield higher immediate loss reduction per unit cost.

To mitigate noisy fluctuations and adapt to the inherently non-stationary nature of training rewards, we maintain an exponential moving average of the reward for each dataset:
\begin{equation}
\label{eq:ema}
    Q_{\mathcal{D}_i} \gets \alpha \cdot Q_{\mathcal{D}_i} + (1-\alpha) \cdot r_{\mathcal{D}_i},
\end{equation}
where $\alpha \in (0,1)$ is the smoothing factor.
This smoothing helps to prevent overreacting to sudden batch-level spikes and ensures that the bandit scheduler remains robust to short-term variance while still tracking long-term trends in dataset utility.

\section{Experimental Setup}
\label{sec:exp_setup}
\subsection{Training Setting}
We conduct experiments with two base models: LLaMA3.2 1B~\citep{llama3.2} and LLaMA3.1 8B~\citep{grattafiori2024llama3herdmodels}. For training data, we use the \tulu-2~\citep{ivison2023camelschangingclimateenhancing} and \tulu-3~\citep{lambert2025tulu3pushingfrontiers} collections, which consist of $\sim$320K examples from 16 datasets and about $\sim$930K examples from 19 datasets, respectively.
Further dataset compositions and implementation details are provided in~\Cref{appendix:dataset_details} and~\ref{appendix:further_implementation_details}.


\subsection{Evaluation}
Following the setup from \citet{lambert2025tulu3pushingfrontiers}, we assess the model on 10 benchmarks covering diverse domain; \textbf{Knowledge}: MMLU~\citep{hendrycks2021measuring}, TruthfulQA~\citep{lin-etal-2022-truthfulqa}, PopQA~\citep{mallen-etal-2023-trust}, \textbf{Reasoning}: BigBench-Hard~\citep{suzgun2022challengingbigbenchtaskschainofthought}, DROP~\citep{dua-etal-2019-drop}, \textbf{Coding}: HumanEval~\citep{chen2021evaluatinglargelanguagemodels}, HumanEval+~\citep{liu2023is}, \textbf{Mathmetical}: GSM8K~\citep{cobbe2021trainingverifierssolvemath}, MATH~\citep{hendrycks2021measuring}, \textbf{Instruction Following}: IFEval~\citep{zhou2023instructionfollowingevaluationlargelanguage}. Further evaluation details are provided in the~\Cref{appendix:evaluation_details}.

\subsection{Baselines}
We compare our method with two baseline groups, \textbf{Heuristic Methods}: \fullBaseline (Proportional sampling) and \uniformBaseline;
and \textbf{Dynamic Methods}: \textit{MultiDDS}~\citep{wang-etal-2020-balancing}, \textit{MultiUAT}~\citep{wu-etal-2021-uncertainty} and \textit{HBO}~\citep{wang2025hbohierarchicalbalancingoptimization}, adaptively adjust dataset mixtures based on gradient cosine similarity, uncertainty and difficulties.


\section{Results}
\label{sec:exp_results}

\begin{table*}[t!]
  \centering
  \resizebox{\textwidth}{!}{%
  \begin{tabular}{lcccccccccc|c}
    \toprule
    \multicolumn{1}{c}{\multirow{2.5}{*}{\textbf{Method}}} &
    \multicolumn{3}{c}{\textbf{General Knowledge}} &
    \multicolumn{2}{c}{\textbf{Reasoning}} &
    \multicolumn{2}{c}{\textbf{Coding}} &
    \multicolumn{2}{c}{\textbf{Mathematical}} &
    \multirow{2.5}{*}{\makecell{IFEval}} &
    \multirow{2.5}{*}{\textbf{AVG}} \\
    \cmidrule(lr){2-4}\cmidrule(lr){5-6}\cmidrule(lr){7-8}\cmidrule(lr){9-10}
     & MMLU & TQA & PopQA & BBH & DROP & CHE & CHE$+$ & GSM8K & MATH &  &  \\ \midrule
    \addlinespace[4pt]
    \multicolumn{12}{c}{\textsc{\textbf{\large \tulu 2 Mixture}}}\\
    \multicolumn{12}{l}{\textsc{\textbf{LLaMA3.2 1B}}}\\
    \uniformBaseline                   & 30.28 & 40.78 & 15.68 & 31.53 & 27.88 & 34.81 & 29.42 & 11.69 & 2.74 & 23.55 & 24.84\\
    \fullBaseline                         & \cellcolor{First}32.13 & 41.12 & \cellcolor{Second}16.22 & 32.93 & 28.18 & 37.23 & 35.47 & \cellcolor{Second}15.77 & \cellcolor{Second}2.91 & 24.78 & 26.67\\
    MultiDDS                   & 31.85 & 41.63 & \cellcolor{First}16.25 & 32.79 & 28.36 & 38.13 & 35.74 & 15.55 & 2.87 & 25.32 & 26.85\\
    MultiUAT                   & \cellcolor{Second}32.11 & 41.59 & 16.12 & \cellcolor{Second}33.52 & 28.37 & 37.28 & 34.33 & 14.51 & 2.81 & 26.18 & 26.68\\
    HBO                   & 31.75 & \cellcolor{Second}41.77 & 16.21 & 32.81 & \cellcolor{Second}28.68 & \cellcolor{Second}38.21 & \cellcolor{Second}35.81 & 15.49 & 2.81 & \cellcolor{Second}26.38 & \cellcolor{Second}26.99\\
    \OURS             & 31.91 & \cellcolor{First}42.02 & 16.18 & \cellcolor{First}33.73 & \cellcolor{First}28.75 & \cellcolor{First}41.69 & \cellcolor{First}36.43 & \cellcolor{First}15.89 & \cellcolor{First}3.12 & \cellcolor{First}27.23 & \cellcolor{First}\textbf{27.70}\\
    \arrayrulecolor{black!50}\midrule
    \multicolumn{12}{l}{\textsc{\textbf{LLaMA3.1 8B}}}\\
    \uniformBaseline                   & 58.78 & 46.71 & 19.33 & 53.21 & 59.83 & 60.51 & 59.55 & 55.31 &  13.1 & 40.44 & 46.68\\
    \fullBaseline                         & \cellcolor{First}61.80 & 49.40 & \cellcolor{First}23.30 & 57.10 & 61.70 & 66.90 & 63.10 & 60.40 & 14.00 & 42.30 & 50.00\\
    MultiDDS                  & \cellcolor{Second}59.81 & 51.67 & 22.82 & 57.11 & 63.53 & \cellcolor{Second}68.84 & 63.76 & \cellcolor{Second}63.25 & 14.84 & 44.29 & \cellcolor{Second}50.99\\
    MultiUAT                  & 59.13 & 51.62 & 22.54 & 57.89 & 63.11 & 68.38 & 63.66 & 63.12 & 14.93 & \cellcolor{Second}44.31 & 50.87\\
    HBO                  & 58.33 & \cellcolor{Second}52.32 & 22.21 & \cellcolor{First}59.44 & \cellcolor{Second}63.81 & 67.91 & \cellcolor{Second}64.30 & 62.14 & \cellcolor{Second}15.24 & 44.12 & 50.98\\
    \OURS             & 59.78 & \cellcolor{First}53.30 & \cellcolor{Second}22.97 & \cellcolor{Second}59.33 & \cellcolor{First}63.91 & \cellcolor{First}71.27 & \cellcolor{First}66.49 & \cellcolor{First}65.93 & \cellcolor{First}16.84 & \cellcolor{First}45.92 & \cellcolor{First}\textbf{52.57}\\
    \arrayrulecolor{black}\midrule[0.9pt]
    \addlinespace[4pt]
    \multicolumn{12}{c}{\textsc{\textbf{\large \tulu 3 Mixture}}}\\
    \multicolumn{12}{l}{\textsc{\textbf{LLaMA3.2 1B}}}\\
    \uniformBaseline                   & 30.45 & 36.58 & 17.45 & 30.73 & 27.84 & 39.32 & 35.12 & 21.12 &  7.11 & 40.98 & 28.67\\
    \fullBaseline                         & 31.09 & \cellcolor{Second}37.80 & \cellcolor{Second}18.26 & 32.01 & \cellcolor{First}28.02 & 43.66 & 42.71 & 28.27 & 8.21 & 44.17 & 31.42\\
    MultiDDS                  & \cellcolor{First}32.78 & 37.21 & 17.98 & 31.73 & 27.84 & 43.75 & 42.43 & 28.43 & 7.94 & 45.63 & 31.57\\
    MultiUAT                  & \cellcolor{Second}32.51 & 37.46 & \cellcolor{First}18.28 & 31.87 & 27.21 & \cellcolor{Second}44.32 & \cellcolor{Second}43.15 & 28.11 & 8.13 & 45.33 & 31.64\\
    HBO                  & 30.79 & 37.63 & 18.02 & \cellcolor{Second}32.34 & \cellcolor{Second}27.94 & 43.97 & 42.82 & \cellcolor{Second}28.82 & \cellcolor{Second}8.57 & \cellcolor{Second}45.87 & \cellcolor{Second}31.68\\
    \OURS             & 32.31 & \cellcolor{First}38.53 & 18.23 & \cellcolor{First}33.21 & 27.92 & \cellcolor{First}46.21 & \cellcolor{First}45.81 & \cellcolor{First}31.44 & \cellcolor{First}9.64 & \cellcolor{First}47.41 & \cellcolor{First}\textbf{33.07}\\
    \arrayrulecolor{black!50}\midrule
    \multicolumn{12}{l}{\textsc{\textbf{LLaMA3.1 8B}}}\\
    \uniformBaseline                   & 60.31 & 45.24 & 28.48& 61.31 & 60.82 & 78.32 & 76.93 & 70.67 &  24.70 & 60.46 & 56.72\\
    \fullBaseline                         & \cellcolor{First}62.10 & 46.80 & 29.30 & 67.90 & 61.30 & 86.20 & 81.40 & 76.20 & 31.50 & 72.80 & 61.55\\
    MultiDDS                  & 61.21 & 45.87 & \cellcolor{Second}29.33 & 68.21 & 61.82 & 87.14 & 82.41 & \cellcolor{Second}78.21 & 30.86 & 71.55 & 61.66\\
    MultiUAT                  & 61.24 & \cellcolor{First}47.14 & 29.11 & 68.18 & 60.12 & \cellcolor{Second}88.23 & 82.13 & 77.91 & 31.41 & 70.13 & 61.56\\
    HBO                  & \cellcolor{Second}61.91 & 46.53 & 28.58 & \cellcolor{Second}68.32 & \cellcolor{First}62.52 & 88.11 & \cellcolor{Second}83.08 & 77.18 & \cellcolor{Second}32.53 & \cellcolor{Second}73.14 & \cellcolor{Second}62.19\\
    \OURS             & 61.87 & \cellcolor{Second}47.12 & \cellcolor{First}29.51 & \cellcolor{First}69.39 & \cellcolor{Second}62.47 & \cellcolor{First}89.14 & \cellcolor{First}84.12 & \cellcolor{First}79.46 & \cellcolor{First}33.21 & \cellcolor{First}74.81 & \cellcolor{First}\textbf{63.11}\\
    \arrayrulecolor{black}\bottomrule
  \end{tabular}
  }
  \caption{Performance comparison across 10 evaluation suites. Best results are shown in dark red and second-best results in light red.}
  \label{tab:main-results}
\end{table*}

\begin{figure*}[t]
  \centering
  \centerline{\includegraphics[width=1.\linewidth]{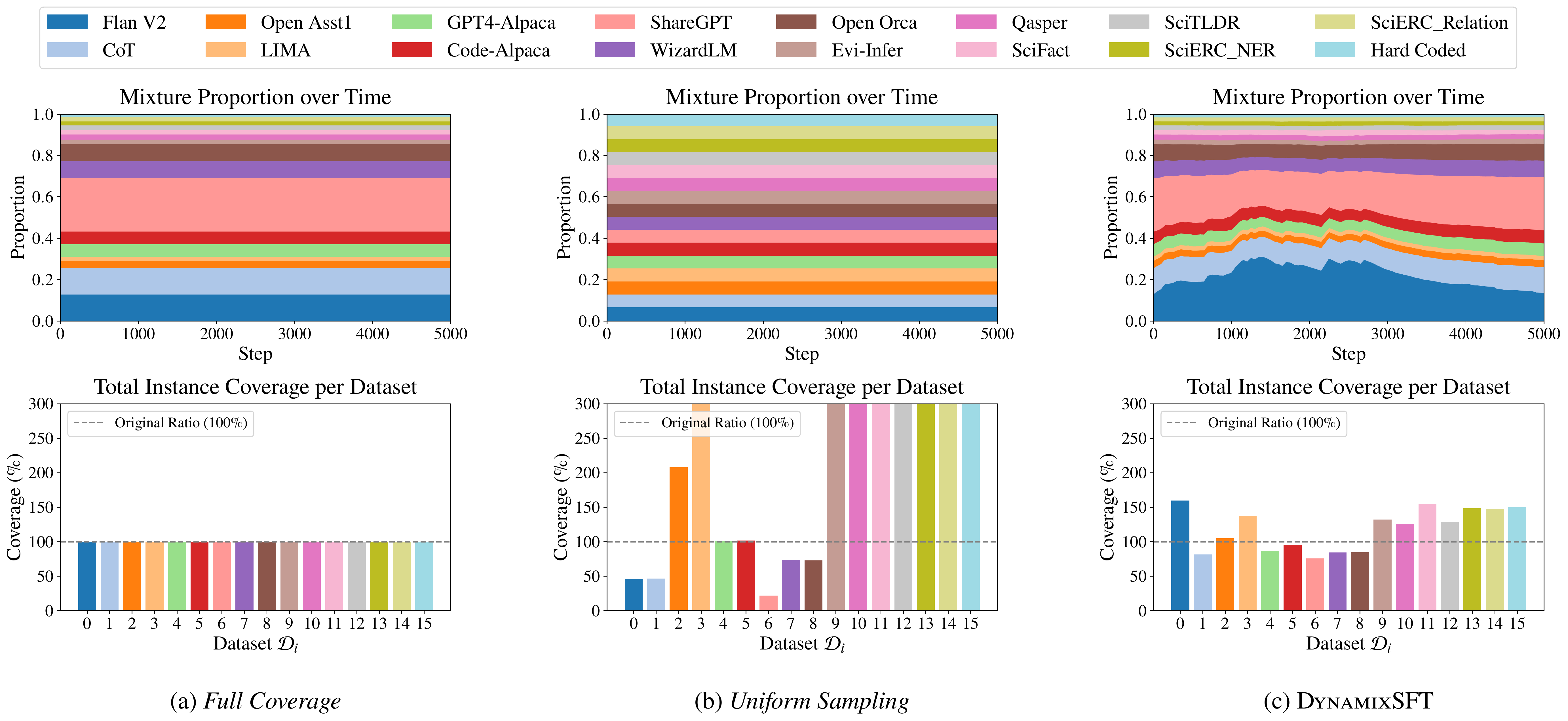}}

  \caption{Visualization of Mixture Proportions and Instance Coverage under different sampling strategies. (a) \textit{Full Coverage}: Mixture proportions are skewed based on dataset sizes (top), but all instances  from each dataset are guaranteed to be used during training (bottom). (b) \textit{Uniform Sampling}: Each dataset is sampled equally (top), leading to severe over- and under-sampling depending on dataset size (bottom). (c) \textsc{DynamixSFT}: Our method adaptively adjusts mixture proportions over time (top), achieving well-balanced instance coverage across all datasets (bottom). All results are based on \mixtureTwo.}
  \label{fig:proportion_coverage}
\end{figure*}

\subsection{Overall Comparison}
In~\Cref{tab:main-results}, \OURS consistently outperforms other baselines across both 1B and 8B models. Compared to naive proportional sampling (\fullBaseline), \OURS achieves relative improvements of up to +5.1\% on \tulu-2 and +5.3\% on \tulu-3 when averaged across 10 benchmarks (see \Cref{appendix:further_implementation_details} for statistical details). Furthermore, \OURS outperforms existing dynamic mixture methods such as HBO~\citep{wang2025hbohierarchicalbalancingoptimization}, which requires an additional actor network, confirming the effectiveness of our lightweight bandit approach.

\subsection{Impact of Instance Coverage}
Among the baselines, the \fullBaseline (proportional sampling) approach shows relatively strong performance for both 1B and 8B models. In contrast, \uniformBaseline consistently underperforms. As illustrated in~\Cref{fig:proportion_coverage}(b), \uniformBaseline disregards the inherent dataset size differences, leading to over-sampling or under-sampling of datasets. This imbalance often distorts the natural characteristics of the original dataset collection, resulting in suboptimal model performance.


\subsection{Visualization of Dynamic Mixture}
\Cref{fig:proportion_coverage} provides an intuitive visualization of how different sampling strategies shape the dataset composition during training. Compared to baselines with static ratios such as \fullBaseline and \uniformBaseline, \OURS dynamically adjusts the mixture weights over time. 
This dynamic behavior is driven by our \textit{\lookaheadReward}, which adapts to the current gradient signal and naturally follows the learning rate schedule. As a result, the sampling proportions become more responsive until mid-training steps, when the learning rate is higher. This adaptivity ultimately produces a more balanced final dataset coverage, as shown in~\Cref{fig:proportion_coverage}(c).

\section{Analysis}
\label{sec:analysis}
\begin{figure}[t]
\begin{center}
\centerline{\includegraphics[width=0.84\columnwidth]{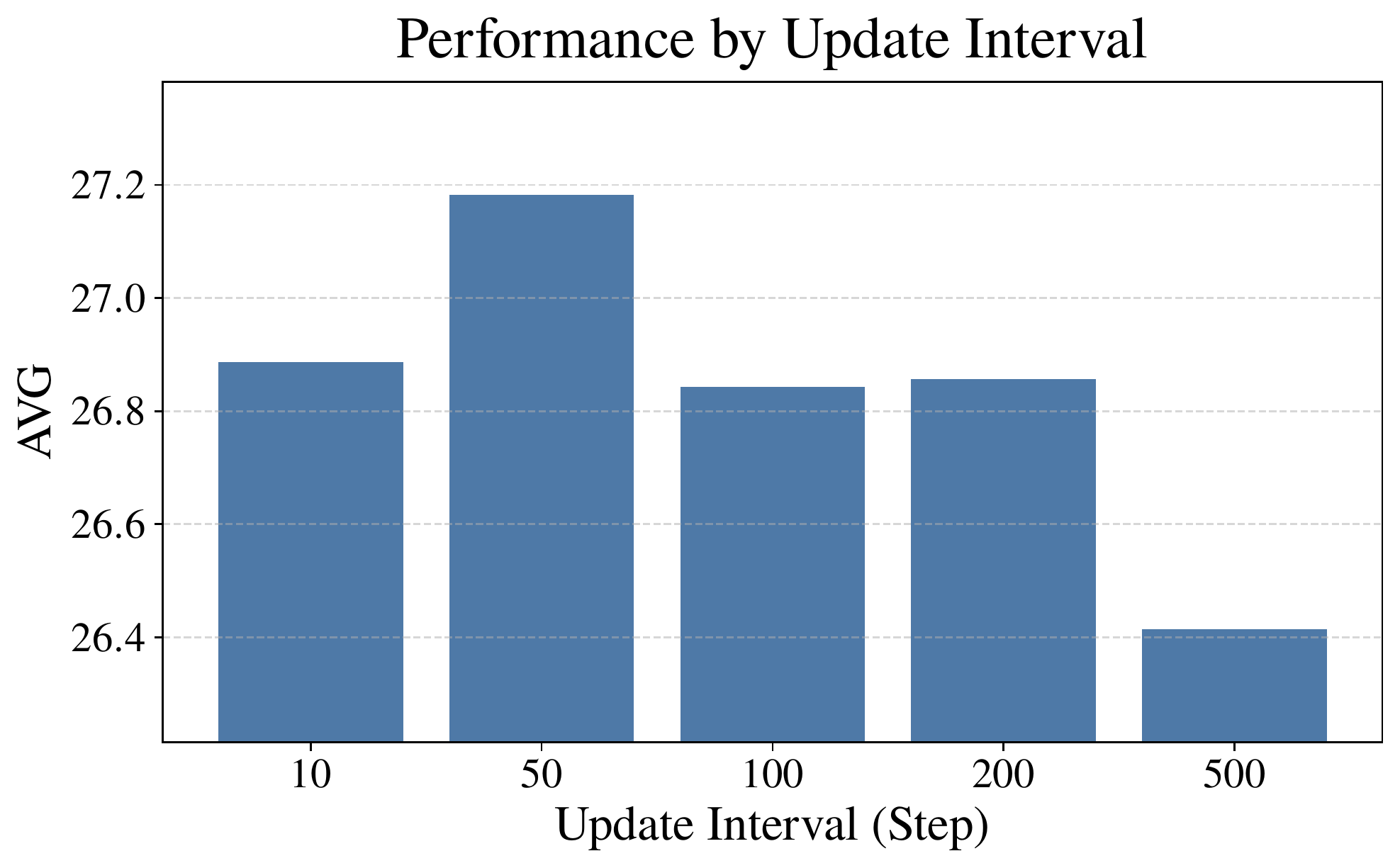}}
\caption{Comparison of Performance by Varying Reward Update Interval.}
\label{fig:update_interval}
\end{center}
\end{figure}

\begin{figure}[t]
\begin{center}
\centerline{\includegraphics[width=1.\columnwidth]{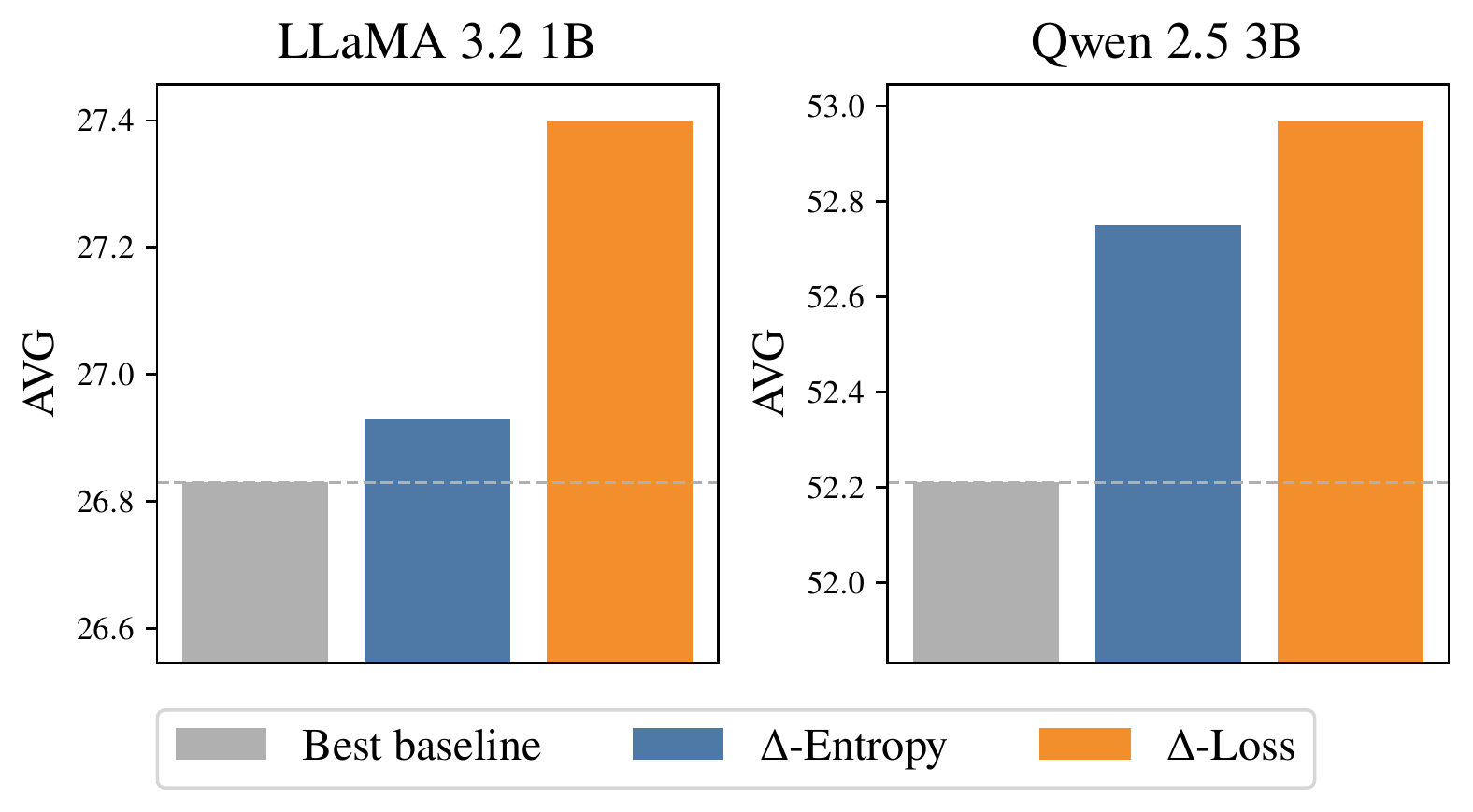}}
\caption{Comparison of Performance Using $\Delta$-Loss vs. $\Delta$-Entropy Rewards.}
\label{fig:loss_entropy}
\end{center}
\end{figure}

\begin{figure}[t]
  \centering
  \subfloat[][As $\gamma$ decreases (blue → orange), the best performance shifts toward higher $\beta$ values, highlighting the trade-off between exploration and exploitation. Extremely large or small values of $\gamma$ (solid lines) exhibit greater performance variance across different $\beta$ values.  \label{subfig:beta_gamma_graph}]{\includegraphics[width=1.\linewidth]{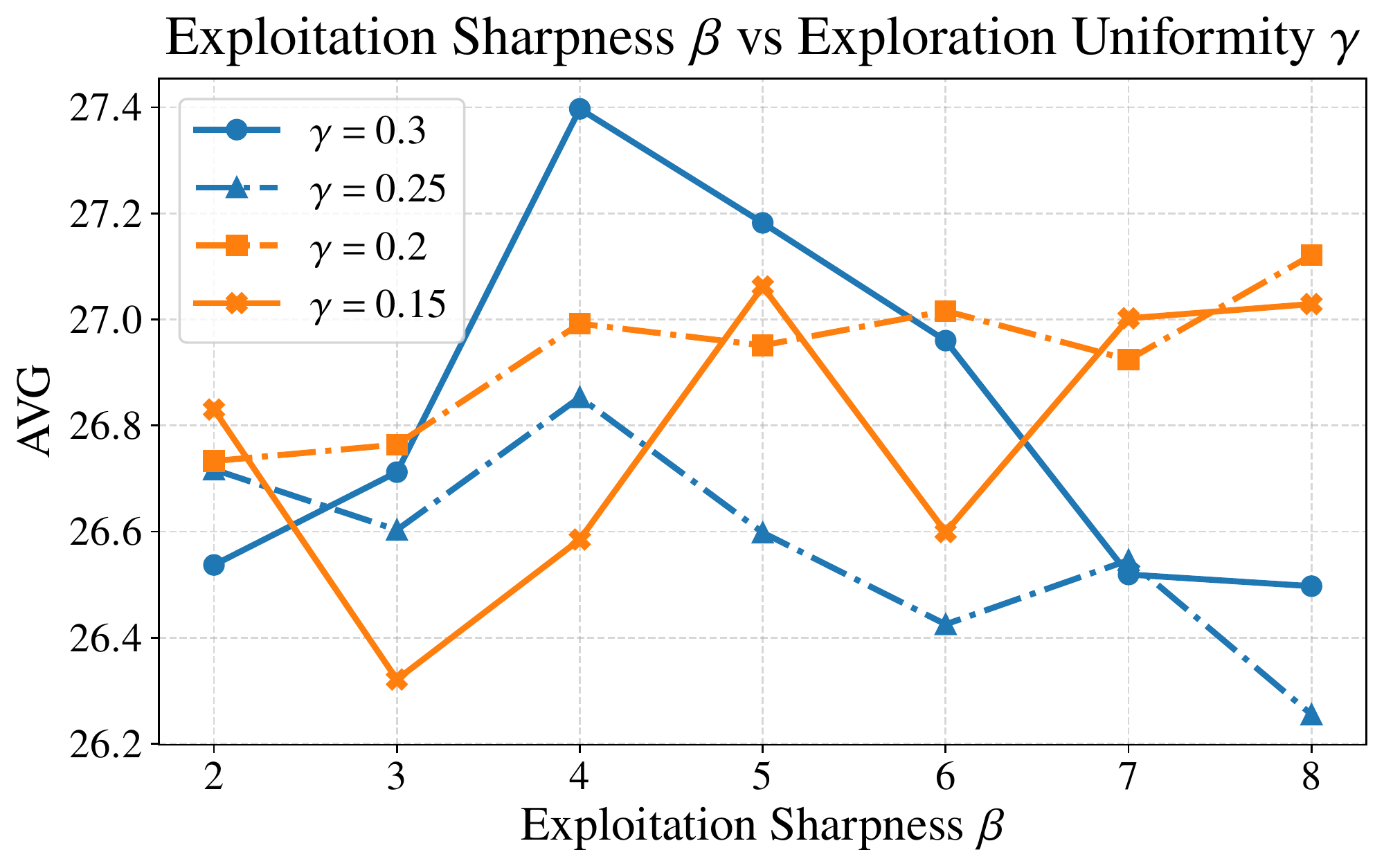}}\hfill
  
  \subfloat[][Extremely large or small values of $\gamma$ show a wide performance variance, as indicated by the spread from blue to red. This suggests that the effect of $\gamma$ can be offset by selecting an appropriate $\beta$. \label{subfig:heatmap}]{\includegraphics[width=1.\linewidth]{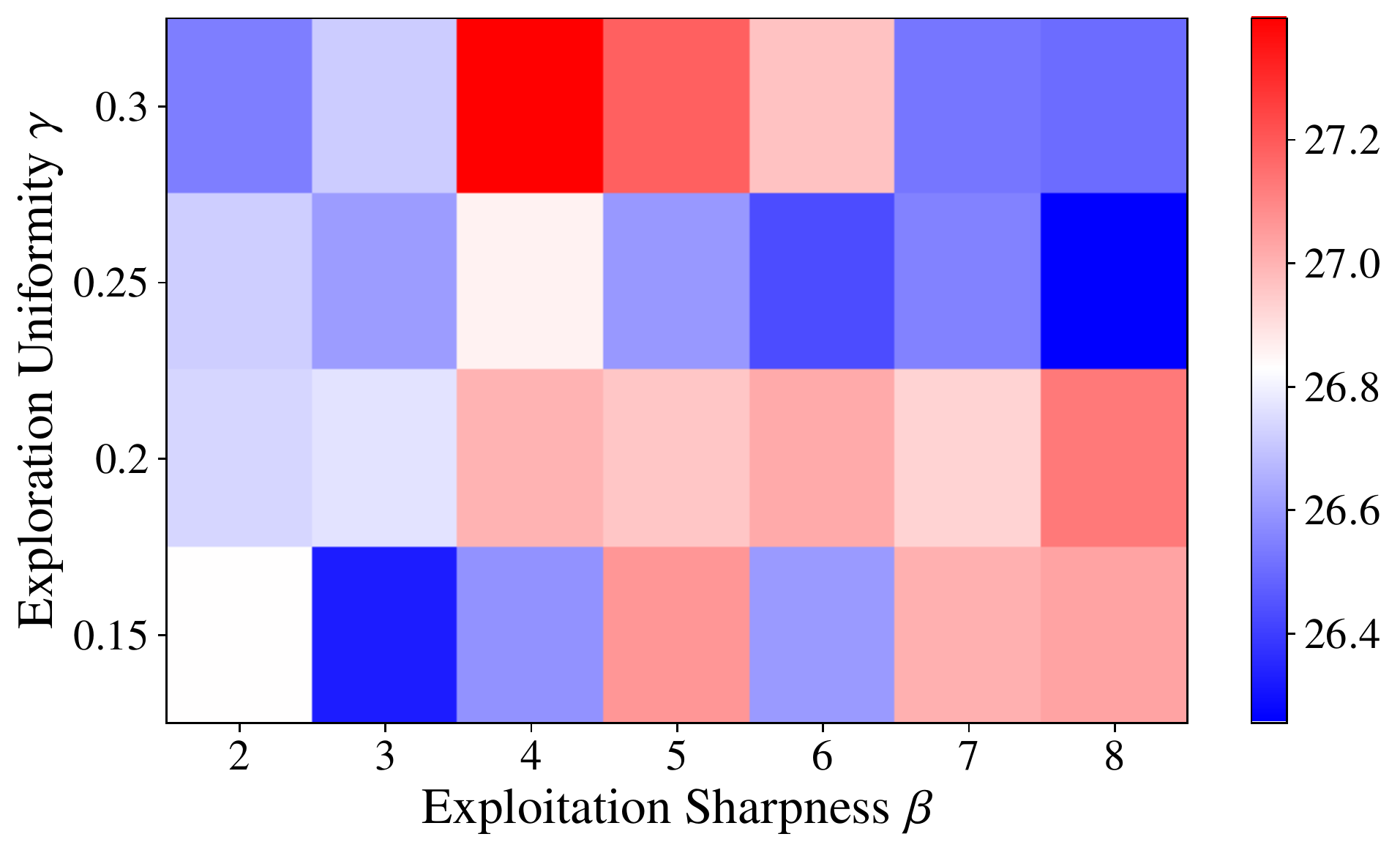}}

  \caption{Performance Comparison varying Exploitation Sharpness $\beta$ vs. Exploration Uniformity $\gamma$. Best viewed in color.}
  \label{fig:exploitation_exploration}
\end{figure}

  


\begin{figure}[t]
  \centering
  \begin{center}
    \subfloat[][Stronger smoothing (darker colors) tends to favor sharper $\beta$ values, while weaker smoothing shifts the optimal $\beta$ to lower values (lighter colors). This suggests that moderate smoothing factor (solid lines) effectively modulates reward sharpness, enabling more stable exploitation behavior. \label{subfig:alpha_beta_graph}]{\includegraphics[width=1.\linewidth]{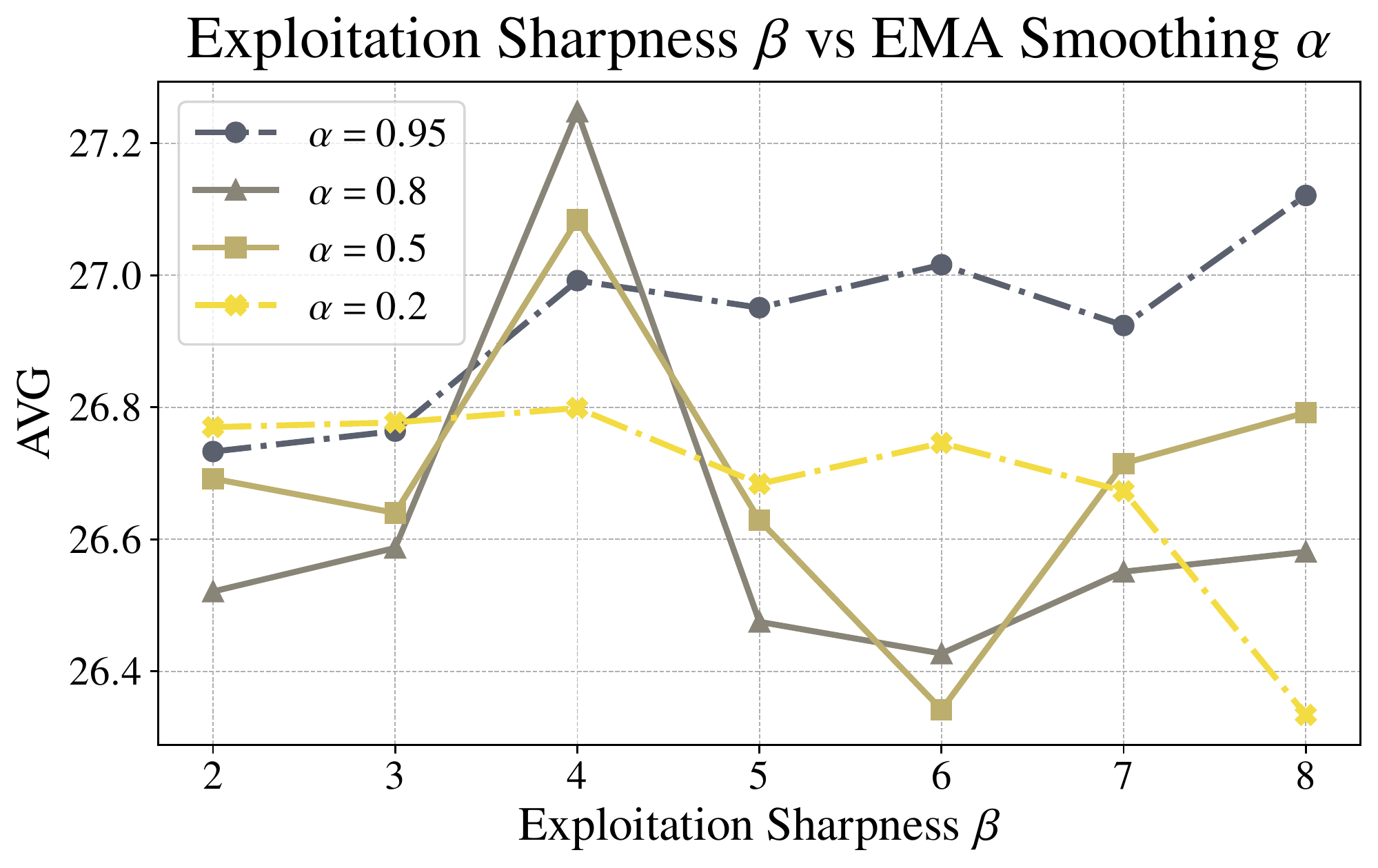}}\hfill

    \subfloat[][As $\alpha$ decreases, the mixture becomes more sensitive to reward signals, resulting in unstable and noisy dynamics. \label{subfig:alpha_4fig}]{\includegraphics[width=1.\linewidth]{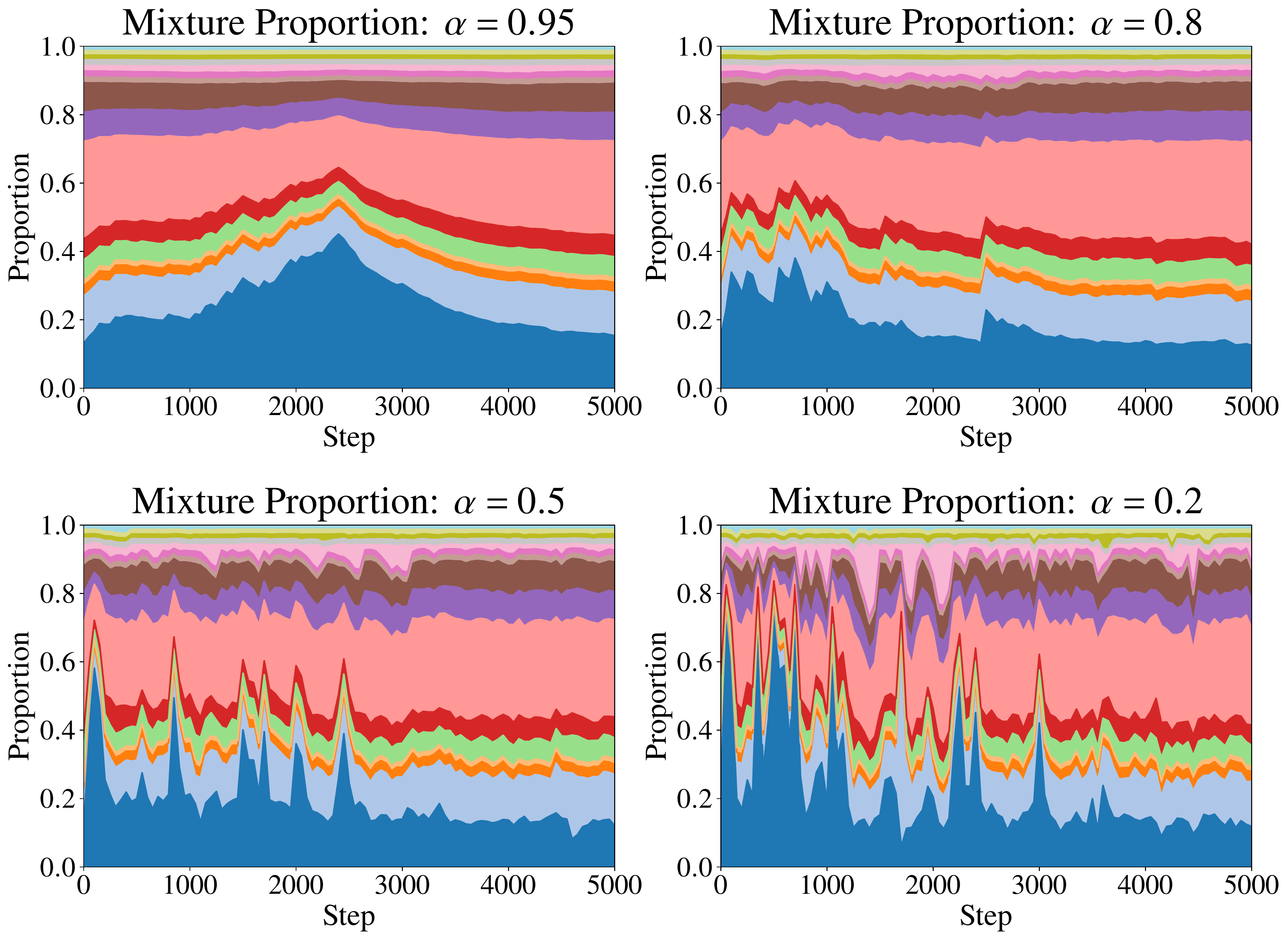}}\hfill


  \caption{Effect of EMA Smoothing Strength $\alpha$ on Performance Trends. Best viewed in color.}
  \label{fig:ema_strength}
  \end{center}
\end{figure}


\begin{table}[t]
  \centering
  \begin{tabular}{lc}
    \toprule
    \textbf{Ablations} & \textbf{AVG} \\
    \midrule
    \OURS                   & 27.7  \\
    \quad \text{w/o prior}      & 25.9  \\
    \quad \text{w/o prior} + prop. init      & 26.3  \\
    \arrayrulecolor{black}\bottomrule
  \end{tabular}
  \caption{Ablation study for the effect of prior distribution $p^{(0)}$ in \Cref{eq:prior_scaled}. \textsc{w/o prior + prop. init} denotes \textsc{w/o prior} with proportional initialization.}
  \label{tab:prior_ablation}
\end{table}

\begin{figure}[t]
  \centering
  \subfloat[][\OURS \label{subfig:proportion_ours}]{\includegraphics[width=0.5\linewidth]{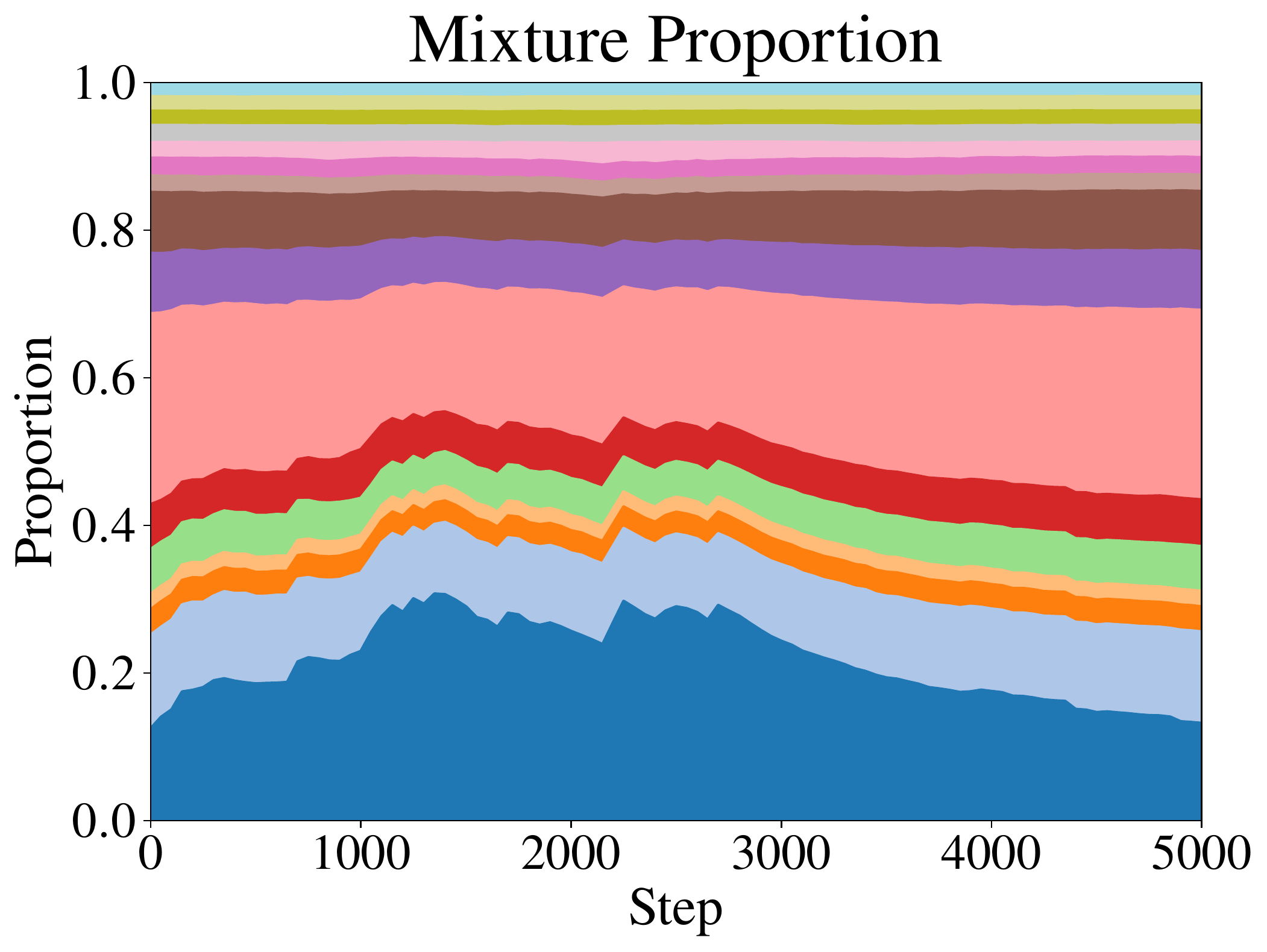}}\hfill
  \subfloat[][\textsc{w/o Prior} \label{subfig:proportion_uniforminit}]{\includegraphics[width=0.5\linewidth]{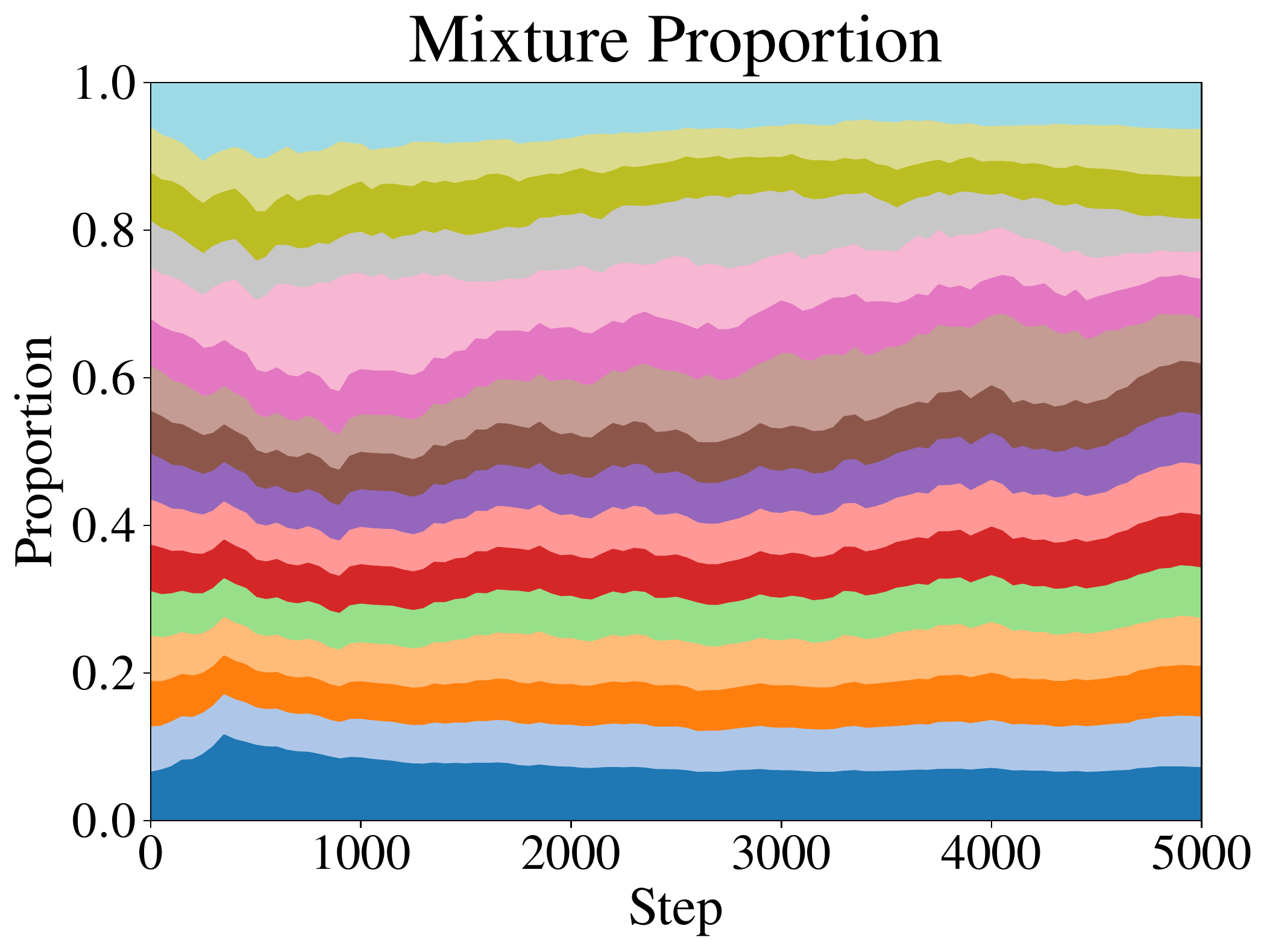}}

  \caption{Effect of removing the prior in \OURS.}
  \label{fig:ablation_wo_prior}
\end{figure}

\subsection{Exploitation vs. Exploration}
In~\Cref{eq:bandit_probs}, we control the exploitation strength using the sharpness factor $\beta$ and the exploration strength using the uniformity factor $\gamma$. As shown in~\Cref{subfig:beta_gamma_graph}, we investigate the effect of varying these two parameters on the 1B model's performance. 

Under strong exploration settings ($\gamma=0.3, 0.25$; blue lines), performance peaks at lower exploitation sharpness values ($\beta=4$). In contrast, under weaker exploration settings ($\gamma=0.2, 0.15$; orange lines), higher sharpness values ($\beta=7, 8$) lead to better performance. Notably, extreme exploration values ($\gamma=0.3, 0.15$; solid lines) lead to greater performance variance across different $\beta$. 
\Cref{subfig:heatmap} visualizes this interaction via a heatmap, where the color gradient (red to blue) indicates significant score variation.
These results highlight the importance of carefully balancing exploitation sharpness and exploration uniformity to achieve optimal mixture performance.

\subsection{Smoothing EMA Reward Strength}
As described in~\Cref{eq:ema}, we control the strength of the exponential moving average reward signal using the smoothing factor $\alpha$, which determines how strongly recent rewards influence the sampling policy. We investigate the effect of different smoothing factor $\alpha$ on performance and mixture dynamics.

In~\Cref{subfig:alpha_beta_graph},  moderate smoothing values ($\alpha=0.8, 0.5$; solid lines) lead to the highest performance when the sharpness factor $\beta=4$, indicating that a balance between responsiveness and stability leads to effective mixture updates. Strong smoothing ($\alpha=0.95$) maintains stable performance even as $\beta$ increases, suggesting that higher exploitation sharpness can be tolerated when updates are sufficiently smoothed. In contrast, weak smoothing ($\alpha=0.2$) causes performance to degrade rapidly as $\beta$ increases, highlighting its inability to regulate over-sharp exploitation.

To better understand these behaviors, we visualize the dynamics of mixture proportions in~\Cref{subfig:alpha_4fig} under a fixed sharpness $\beta = 4$. Strong smoothing ($\alpha=0.95$) produces smooth and gradually evolving proportions, while weaker smoothing ($\alpha=0.2$) leads to highly volatile and unstable allocation, frequently switching focus among datasets. These results highlight the importance of EMA smoothing for stable and robust mixture learning.


\subsection{Varying Update Interval}
We evaluate how different update intervals affect overall performance. As shown in~\Cref{fig:update_interval}, updating the sampling policy every 50 steps yields the best performance. While shorter intervals (e.g., 10 steps) enable quicker adaptation to recent rewards, they may introduce instability. Conversely, longer intervals (e.g., 500 steps) lead to slower adaptation and slightly degraded performance. These findings suggest that moderately frequent updates strike a good balance between responsiveness and stability in mixture optimization.

\subsection{Reward Comparison with $\Delta$-Loss vs. $\Delta$-Entropy}
To further explore the potential of the \textit{\lookaheadReward}, we investigate an alternative reward formulation based on entropy difference. As shown in~\Cref{fig:loss_entropy}, using the entropy-based reward ($\Delta$-Entropy) also yields performance gains over the baselines in both 1B and 8B. Notably, for the 8B model, $\Delta$-Entropy achieves performance comparable to the $\Delta$-Loss, suggesting that uncertainty reduction can serve as an effective signal for guiding mixture optimization. These results demonstrate the flexibility of our framework in accommodating different reward definitions while still improving overall performance.

\subsection{Effect of Prior Scaling}
We analyze how the prior-scaled formulation $p^{(0)}$ in \Cref{eq:prior_scaled} affects overall performance on the 
\mixtureTwo with a 1B model. \Cref{tab:prior_ablation} shows that removing the $p^{(0)}$ (\textsc{w/o prior} in~\Cref{tab:prior_ablation}) leads to a substantial performance drop. To better understand this effect, we compare the mixture dynamics with and without $p^{(0)}$ in~\Cref{fig:ablation_wo_prior}. Without prior-scaling, the sampling distribution remains close to its initial uniform distribution throughout training, showing that model fails to discover a good mixture.

Even when we manually set the initial distribution to be proportional (\textsc{w/o prior + prop. init} in~\Cref{tab:prior_ablation}), a performance gap still remains compared to the prior-scaled formulation. This demonstrates that prior-scaling not only naturally initializes the mixture near the data's intrinsic distribution, but also continuously anchors it throughout training, preventing drift during reward-based optimization.


\section{Efficiency}
\label{sec:efficiency}
\begin{table}[t]
\centering
\begin{tabular}{lr}
\toprule
\textbf{Method} & \textbf{Overhead vs. Naive} \\
\midrule
MultiDDS  & +760\% (8.6$\times$)   \\
MultiUAT  & +380\% (4.8$\times$)   \\
HBO       & +139\% (2.39$\times$) \\
\OURS     & +12.7\% (1.13$\times$)  \\
\bottomrule
\end{tabular}
\caption{Computational overhead of dynamic data mixture methods relative to naive sampling.}
\label{tab:overhead}
\end{table}

We analyze the computational overhead of \OURS against existing dynamic data mixture approaches, under the \mixtureThree setup with $K=19$ datasets and $T_{\text{update}}=50$. In~\Cref{tab:overhead}, \OURS introduces only $\sim$12.7\% additional training cost over naive sampling, while prior methods incur substantially higher overhead (+139\% to +760\%). 

This efficiency stems from our \lookaheadReward, which relies solely on forward passes and requires no additional backward computation. Assuming the standard ratio where a forward pass accounts for approximately 1/3 of a full training step, the theoretical overhead is $\frac{19}{50} \times \frac{1}{3} \approx 12.7\%$.

In contrast, existing approaches incur substantially higher overhead, as they rely on gradient computations across all datasets~\citep{wang-etal-2020-balancing, wang2025hbohierarchicalbalancingoptimization} or repeated stochastic forward passes for Monte Carlo estimation~\citep{wu-etal-2021-uncertainty}.
Overall, \OURS achieves strong performance while maintaining minimal computational overhead, making it well-suited for large-scale data mixture training.

\section{Conclusion}
\label{sec:conclusion}
We propose \OURS, a method for dynamic mixture optimization of instruction-tuning collections. We formulate the dataset mixture problem as a multi-armed bandit setup and introduce a \textit{\priorScaled} to softly anchor the sampling distribution. Each dataset is treated as an arm and updated using a lightweight \textit{\lookaheadReward}, without requiring additional models or a validation set. We demonstrate the effectiveness of \OURS on the \mixtureTwo and \mixtureThree collections, without relying on human-crafted criteria. Furthermore, we provide extensive analyses and visualizations that illustrate how \OURS adapts mixture dynamics over time and highlights the contribution of each dataset during training.

\section*{Limitations}
\label{sec:limitations}
In this work, we focus dataset mixture optimization in instruction-tuning stage, where the dataset collection consists of large-scale, heterogeneous datasets aggregated without explicit domain structuring.
Instead of conducting fine-grained analysis on individual datasets or domain-specific effects, our focus lies in optimizing mixture quality at scale across diverse sources.
Moreover, our mixture policy operates at the dataset level, assigning sampling weights to entire datasets rather than individual instances. While this design enables efficient optimization and interpretable mixture control, instance-level mixture strategies remain an open direction for future work.



\section*{Ethical Considerations}
\label{sec:limitations}
This work employs the \tulu collections, which is distributed under the ODC-BY license. The dataset combines multiple publicly available and web-sourced subsets, some of which are subject to non-commercial or mixed-use restrictions. All experiments in this study were conducted strictly for academic research purposes without any commercial intent.
Given the inclusion of web-crawled data, a small portion of the corpus may inadvertently contain imperfect or biased content. We follow the data usage policies provided by the dataset maintainers and do not perform any additional data redistribution or manual modification. All experiments strictly comply with the dataset's research-only usage guidelines.

\section*{Acknowledgements}
\label{sec:acknowledgements}
This research was supported by the MSIT (Ministry of Science, ICT), Korea, under the Global Research Support Program in the Digital Field program (RS-2024-00436680) supervised by the IITP (Institute for Information \& Communications Technology Planning \& Evaluation). And, this project was also supported by Microsoft Research Asia.

\bibliography{custom}

\clearpage
\newpage
\appendix

\label{sec:appendix}
\section{Implementation Details}
\label{appendix:further_implementation_details}
Following the \tulu training setup~\citep{ivison2023camelschangingclimateenhancing, lambert2025tulu3pushingfrontiers}, we train the model for 2 epochs, using a batch size of 128 for both training and reward updates. All training is conducted on 8 $\times$ A100 40GB GPUs. In accordance with the each dataset collection, the number of arms $K$ is set to 16 for \tulu-2 and 19 for \tulu-3.
We report results using $\gamma = 0.3$, $\alpha = 0.95$, and $\beta = 4$ for the 1B model, and $\beta = 5$ for the 8B model. The learning rate is set to 1e-5, with a linear decay scheduler and a warmup ratio of 0.03. 
Reported results are averaged over three runs. The standard deviations for \OURS were consistently small ($\sigma < 0.25$), and a t-test against the primary baseline yielded $\text{p-value} < 0.05$, confirming the statistical significance of the improvements.




\section{Details of Evaluation Suite}
\label{appendix:evaluation_details}
We evaluate our model on 10 benchmarks, following the \tulu-3 evaluation setup~\citep{lambert2025tulu3pushingfrontiers}, as detailed in~\Cref{tab:evaluation_details}.

\begin{table}[h!]
  \centering
  \resizebox{1.0\linewidth}{!}{%
  \begin{tabular}{lcccc}
    \toprule
    Benchmark & CoT & Shots & Multiturn ICL & Metric \\
    \midrule
    MMLU & \cmark & 0 & \xmark & EM \\
    TQA & \xmark & 6 & \xmark & MC2 \\
    PopQA & \xmark & 15 & \cmark & EM \\
    \arrayrulecolor{black!30}\midrule
    BBH & \cmark & 3 & \cmark & EM \\
    DROP & \xmark & 3 & \xmark & F1 \\
    \arrayrulecolor{black!30}\midrule
    CHE & \xmark & 0 & \xmark & Pass@10 \\
    CHE$+$ & \xmark & 0 & \xmark & Pass@10 \\
    \arrayrulecolor{black!30}\midrule
    GSM8K & \cmark & 8 & \cmark & EM \\
    MATH & \cmark & 4 & \cmark & Flex EM \\
    \arrayrulecolor{black!30}\midrule
    IFEval & \xmark & 0 & \xmark & Pass@1 \\
    \arrayrulecolor{black}\bottomrule
  \end{tabular}
  }
  \caption{Evaluation Setup for \OURS.}
  \label{tab:evaluation_details}
\end{table}

\section{Details of Dataset Collection}
\label{appendix:dataset_details}
We utilize the \mixtureTwo and \mixtureThree collections.
\mixtureTwo comprises $\sim$320K instances from 16 instruction-tuning datasets: FLAN (50K)~\citep{longpre2023flancollectiondesigningdata}, FLAN-CoT (50K)~\citep{longpre2023flancollectiondesigningdata}, Open Assistant 1 (7K)~\citep{NEURIPS2023_949f0f8f}, ShareGPT (114K), GPT4-Alpaca (20K)~\citep{peng2023instructiontuninggpt4}, Code Alpaca (20K)~\citep{codealpaca}, LIMA (1K)~\citep{NEURIPS2023_ac662d74}, WizardLM (30K)~\citep{xu2024wizardlm}, Open-Orca (30K)~\citep{OpenOrca}, Hardcoded (140)~\citep{ivison2023camelschangingclimateenhancing}, and a science-related dataset (7K) combining Evidence Inference~\citep{lehman-etal-2019-inferring}, Qasper~\citep{dasigi-etal-2021-dataset}, SciERC~\citep{luan-etal-2018-multi}, SciFact~\citep{wadden-etal-2020-fact}, and SciTLDR~\citep{cachola-etal-2020-tldr}.

\mixtureThree contains $\sim$930K instances from 19 datasets: 
CoCoNot (10K)~\citep{NEURIPS2024_58e79894},
FLAN v2 (89K)~\citep{longpre2023flancollectiondesigningdata},
No Robots (9K)~\citep{no_robots},
OpenAssistant Guanaco (7K)~\citep{NEURIPS2023_949f0f8f},
Tulu 3 Persona MATH (149K),
Tulu 3 Persona GSM (49K),
Tulu 3 Persona Python (34K), 
Tulu 3 Persona Algebra (20K),
Tulu 3 Persona IF (29K),
NuminaMath-TIR (64K)~\citep{numina_math_datasets},
OpenMathInstruct 2 (50K)~\citep{toshniwal2024openmathinstruct2acceleratingaimath},
Tulu 3 WildGuardMix (50K),
Tulu 3 WildJailbreak (50K),
Tulu 3 Hardcoded (240),
Aya (100K)~\citep{singh2024aya},
WildChat GPT-4 (100K)~\citep{zhao2024wildchat},
TableGPT (5K)~\citep{li2024tablegpt},
SciRIFF (10K)~\citep{wadden-etal-2025-sciriff}.
Evol CodeAlpaca (107K)~\citep{luo2023wizardcoder}.

\section{Long-term Effects of Immediate Reward}
\label{appendix:theoretical_support}
We demonstrate that optimizing the \lookaheadReward (immediate objective) is theoretically equivalent to achieving optimal convergence in the long term.

Given a loss function $\mathcal{L}(\theta)$ and standard gradient 
update $\theta_{t+1} = \theta_t - \eta \nabla \mathcal{L}_k(\theta_t)$, 
let's consider the immediate reward:
\begin{equation}
    r_{t,k} = \mathcal{L}(\theta_t) - \mathcal{L}(\theta_{t+1}).
\end{equation}
Assuming $\mathcal{L}$ is smooth and the learning rate $\eta$ is sufficiently small, we can approximate $\mathcal{L}(\theta_{t+1})$ using the first-order Taylor expansion around $\theta_t$:
\begin{equation}
    \mathcal{L}(\theta_{t+1}) \approx \mathcal{L}(\theta_t) - \eta || \nabla \mathcal{L}(\theta_t) ||^2,
\end{equation}
and rearranging the terms to express the reward:
\begin{equation}
    r_{t,k} = \mathcal{L}(\theta_t) - \mathcal{L}(\theta_{t+1}) \approx \eta || \nabla \mathcal{L}(\theta_t) ||^2.
\end{equation}

Selecting dataset $k$ to maximize the 1-step reward $r_{t,k}$ is effectively equivalent to maximizing the descent magnitude along the gradient direction. 
Therefore, our reward does not track raw loss value, it explicitly measures learning progress.

\paragraph{Why MAB is Appropriate?}
The Multi-Armed Bandit (MAB) framework is naturally suited for this task because it optimizes the \textit{cumulative reward} $\sum_{t=0}^T r_t$ over the entire training horizon. 
In our setting, the reward is defined as the immediate loss reduction: $r_t = \mathcal{L}(\theta_t) - \mathcal{L}(\theta_{t+1})$. 
By the telescoping property, the cumulative reward is:
\begin{equation}
    \sum_{t=0}^T r_t = \sum_{t=0}^T [\mathcal{L}(\theta_t) - \mathcal{L}(\theta_{t+1})] = \mathcal{L}(\theta_0) - \mathcal{L}(\theta_{T+1})
\end{equation}
Consequently, using the $\Delta$-Loss reward within a bandit policy is not just a heuristic trick, it is a principled policy that prioritizes datasets yielding greater long-term loss reduction while maintaining essential exploration.




\end{document}